\documentclass[11pt,a4paper]{article}
\usepackage[hyperref]{naaclhlt2019}
\usepackage{times}
\usepackage{latexsym}
\usepackage{graphicx}
\usepackage{subcaption}
\usepackage{tikz}
\usepackage{microtype}
\usepackage{amsfonts}

\usepackage{rotating}
\def\checkmark{\tikz\fill[scale=0.4](0,.35) -- (.25,0) -- (1,.7) -- (.25,.15) -- cycle;}

\usepackage{array,multirow}
\graphicspath{figures}

\usepackage{url}
\usepackage{adjustbox}

\usepackage[disable]{todonotes} 

\aclfinalcopy 
\def\aclpaperid{2117} 

\newcommand{\word}[1]{\textit{#1}}
\newcommand{\saveforCR}[1]{}

\title{Gender Bias in Contextualized Word Embeddings}

\author{Jieyu Zhao$^\S$ \qquad
 Tianlu Wang$^\dagger$ \qquad
  Mark Yatskar$^\ddag$   \\
{\bf Ryan Cotterell$^\aleph$ \qquad Vicente Ordonez$^\dagger$ \qquad
 Kai-Wei Chang$^\S$}
\\
  $^\S$University of California, Los Angeles \qquad
  \{jyzhao, kwchang\}@cs.ucla.edu \\
  $^\dagger$University of Virginia \qquad
  \{tw8bc, vicente\}@virginia.edu
  \\ $^\ddag$Allen Institute for Artificial Intelligence \qquad marky@allenai.org
  \\ $^\aleph$University of Cambridge \qquad rdc42@cam.ac.uk
}

\date{}

\begin{document}
\maketitle
\begin{abstract}
 In this paper, we quantify, analyze and mitigate gender bias exhibited in ELMo's contextualized word vectors.
 First, we conduct several intrinsic analyses and find that (1) training data for ELMo contains significantly more male than female entities, (2) the trained ELMo embeddings systematically encode gender information and (3) ELMo unequally encodes gender information about male and female entities.
 Then, we show that a state-of-the-art coreference system that depends on ELMo inherits its bias and demonstrates significant bias on the WinoBias probing corpus.
 Finally, we explore two methods to mitigate such gender bias and show that the bias demonstrated on WinoBias can be eliminated.

\end{abstract}

\section{Introduction}
\label{sec:introduction}
Distributed representations of words in the form of word embeddings~\cite{mikolov2013distributed,pennington2014glove} and contextualized word embeddings~\cite{Peters:2018,devlin2018bert,radford2018improving,mccann2017learned,radford2019language} have led to huge performance improvement on many NLP tasks.
However, several recent studies show that training word embeddings in large corpora could lead to encoding societal biases present in these human-produced data~\cite{BCWS16,caliskan2017semantics}. In this work, we extend these analyses to the ELMo contextualized word embeddings.

Our work provides a new intrinsic analysis of how ELMo represents gender in biased ways.
First, the corpus used for training ELMo has a significant gender skew: male entities are nearly three times more common than female entities, which leads to gender bias in the downloadable pre-trained contextualized embeddings.
Then, we apply principal component analysis (PCA) to show that after training on such biased corpora, there exists a low-dimensional subspace that captures much of the gender information in the contextualized embeddings.
Finally, we evaluate how faithfully ELMo preserves gender information in sentences by measuring how predictable gender is from ELMo representations of occupation words that co-occur with gender revealing pronouns.
Our results show that ELMo embeddings perform unequally on male and female pronouns: male entities can be predicted from occupation words 14\% more accurately than female entities.\looseness=-1

In addition, we examine how gender bias in ELMo propagates to the downstream applications. Specifically, we evaluate a state-of-the-art coreference resolution system ~\cite{lee2018higher} that makes use of ELMo's contextual embeddings on WinoBias~\cite{ZWYOC18}, a coreference diagnostic dataset that evaluates whether systems behave differently on decisions involving male and female entities of stereotyped or anti-stereotyped occupations.
We find that in the most challenging setting, the ELMo-based system has a disparity in accuracy between pro- and anti-stereotypical predictions, which is nearly 30\% higher than a similar system based on GloVe~\cite{lee2017end}.

Finally, we investigate approaches for mitigating the bias which propagates from the contextualized word embeddings to a coreference resolution system.
We explore two different strategies: (1) a training-time data augmentation technique~\cite{ZWYOC18}, where we augment the corpus for training the coreference system with its gender-swapped variant (female entities are swapped to male entities and vice versa) and, afterwards, retrain the coreference system; and (2) a test-time embedding neutralization technique, where input contextualized word representations are averaged with word representations of a sentence with entities of the opposite gender.
Results show that test-time embedding neutralization is only partially effective, while data augmentation largely mitigates bias demonstrated on WinoBias by the coreference system.\looseness=-1

\section{Related Work}

Gender bias has been shown to affect several real-world applications relying on automatic language analysis, including online news~\cite{ross2011women}, advertisements~\cite{sweeney2013discrimination}, abusive language detection~\cite{park2018reducing}, machine translation~\cite{font2019equalizing,Vanmassenhove2018getting}, and web search~\cite{kay2015unequal}. In many cases, a model not only replicates bias in the training data but also amplifies it~\cite{ZWYOC17}.

For word representations, \newcite{BCWS16} and \newcite{caliskan2017semantics} show that word embeddings encode societal biases about gender roles and occupations, e.g.~engineers are stereotypically men, and nurses are stereotypically women.
As a consequence, downstream applications that use these pretrained word embeddings also reflect this bias.
For example, \newcite{ZWYOC18} and \newcite{rudinger-EtAl:2018:N18} show that coreference resolution systems relying on word embeddings encode such occupational stereotypes.
In concurrent work, \citet{may2019measuring} measure gender bias in sentence embeddings, but their evaluation is on the aggregation of word representations. In contrast, we analyze bias in contextualized word representations and its effect on a downstream task.

To mitigate bias from word embeddings, \newcite{BCWS16} propose a post-processing method to project out the bias subspace from the pre-trained embeddings.
Their method is shown to reduce the gender information from the embeddings of gender-neutral words, and, remarkably, maintains the same level of performance on different downstream NLP tasks.
\newcite{zhao2018learning} further propose a training mechanism to separate gender information from other factors.
However, \newcite{Gonen2019LipstickOA2} argue that entirely removing bias is difficult, if not impossible, and the gender bias information can be often recovered.
This paper investigates a natural follow-up question: What are effective bias mitigation techniques for contextualized embeddings?

\section{Gender Bias in ELMo}
\label{sec:visualizing_bias}
\begin{figure*}[!th]
\centering
    \begin{subfigure}[b]{0.4\textwidth}
        \hspace{-0.7em}\includegraphics[width=1.14\textwidth]{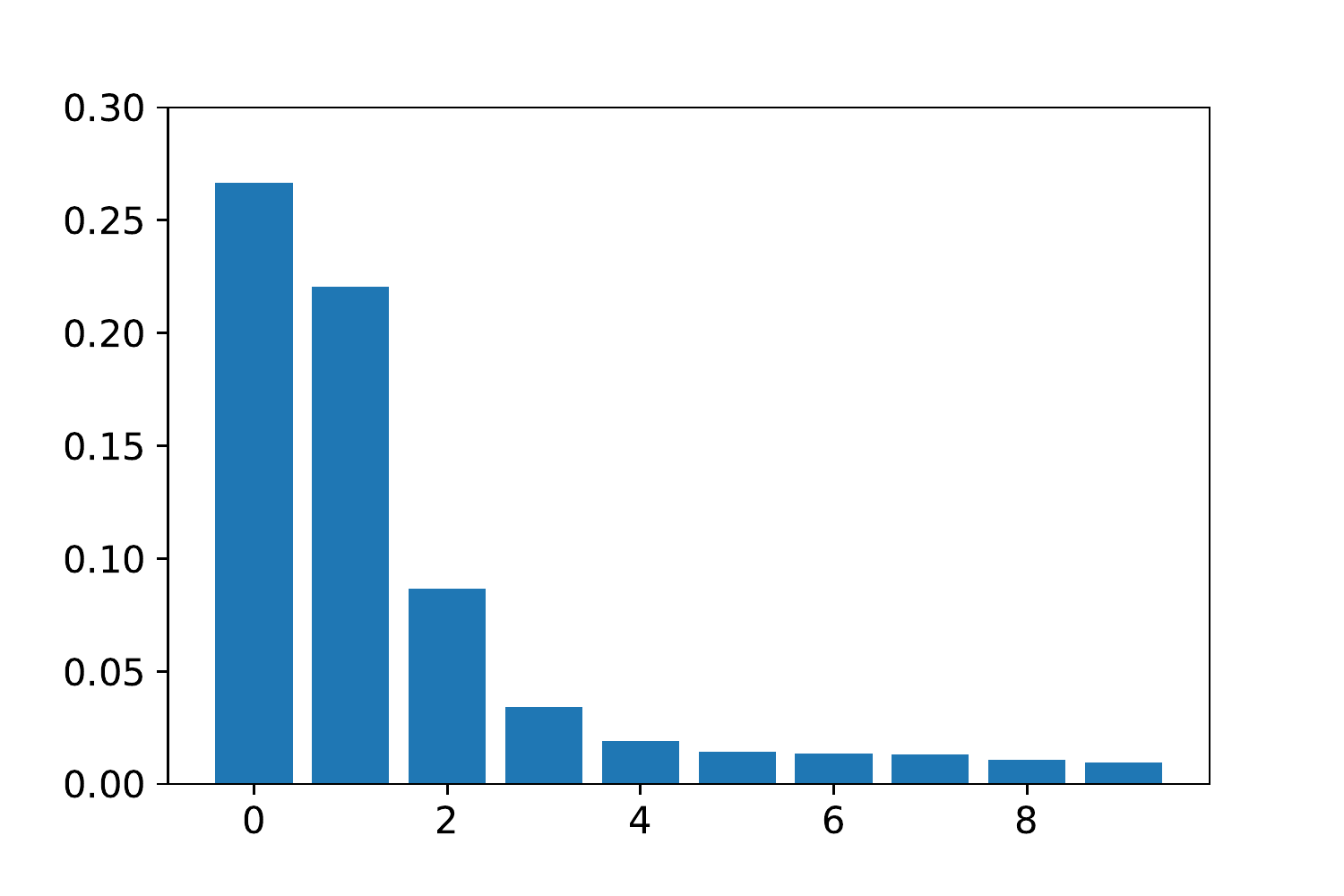}
        \label{fig:pca1}
    \end{subfigure}
    ~
    \begin{subfigure}[b]{0.4\textwidth}
        \hspace{-1em}\includegraphics[width=1.14\textwidth]{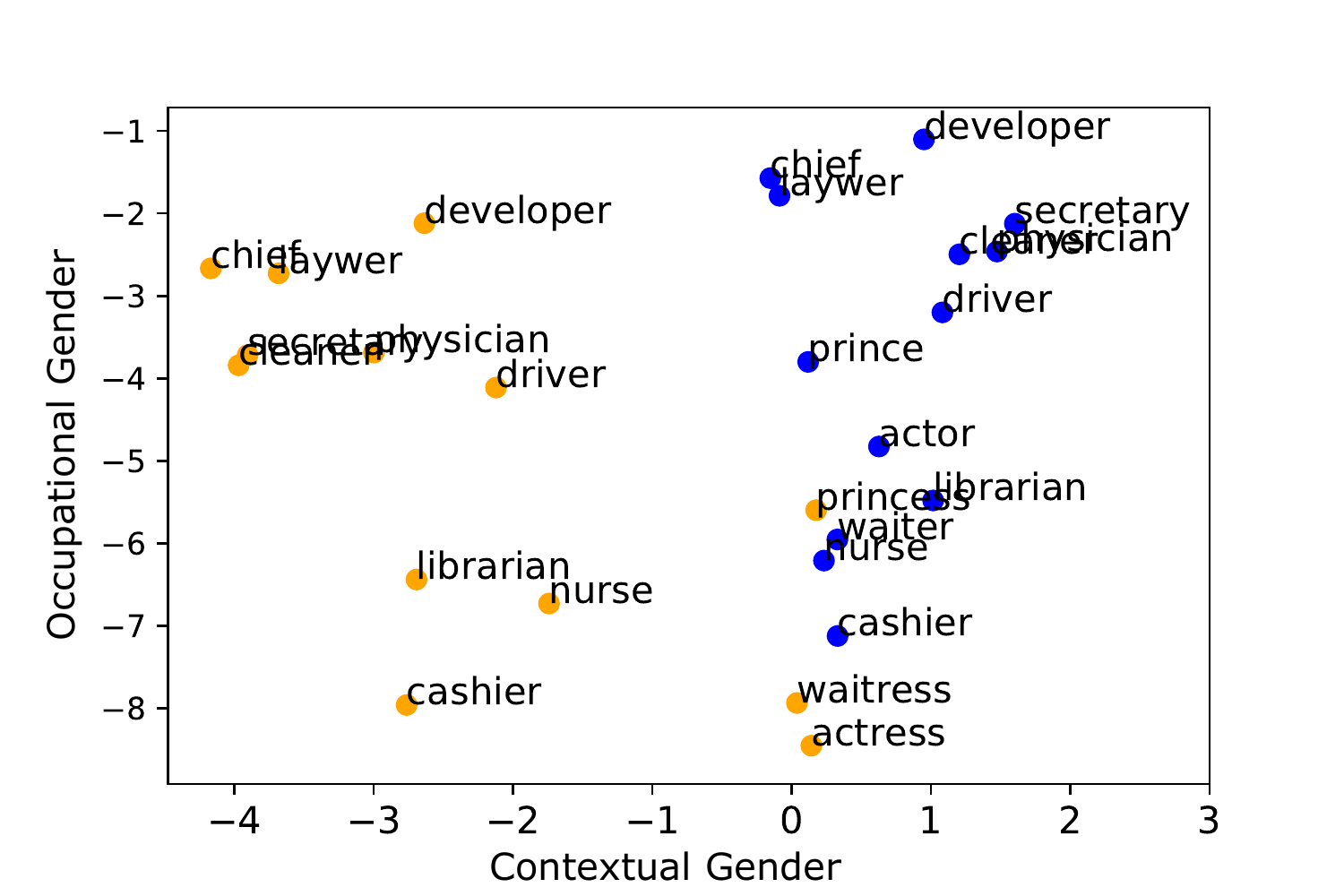}
        \label{fig:pc12}
    \end{subfigure}
    \vspace{-10pt}
    \caption{
    Left: Percentage of explained variance in PCA in the embedding differences.
    Right: Selected words projecting to the first two principle components where the blue dots are the sentences with male context and the orange dots are from the sentences with female context.
    }
    \label{fig:pca}
    \vspace{-10pt}
\end{figure*}

In this section we describe three intrinsic analyses highlighting gender bias in trained ELMo contextual word embeddings~\cite{Peters:2018}.
We show that (1) training data for ELMo contains significantly more male entities compared to female entities leading to gender bias in the pre-trained contextual word embeddings (2) the geometry of trained ELMo embeddings systematically encodes gender information and (3) ELMo propagates gender information about male and female entities unequally.

\subsection{Training Data Bias}
\begin{table}[t]
\small
    \centering
    \begin{adjustbox}{width=\columnwidth}
    \begin{tabular}{c|c|c|c}
    \hline
    & \#occurrence & \#M-biased occs. &\#F-biased occs. \\
    \hline
    M & 5,300,000& 170,000& 81,000 \\
    F & 1,600,000&33,000 &36,000\\
         \hline
    \end{tabular}
     \end{adjustbox}
    \caption{Training corpus for ELMo. We show total counts for male (M) and female (F) pronouns in the corpus, and counts corresponding to their co-occurrence with occupation words where the occupations are stereotypically male (M-biased) or female (F-biased). }
    \vspace{-10pt}
    \label{tab:lm_cor}
\end{table}

Table~\ref{tab:lm_cor} lists the data analysis on the One Billion Word Benchmark~\cite{chelba2013one} corpus, the training corpus for ELMo.
We show counts for the number of occurrences of male pronouns (\word{he}, \word{his} and \word{him}) and female pronouns (\word{she} and \word{her}) in the corpus as well as the co-occurrence of occupation words with those pronouns.
We use the set of occupation words defined in the WinoBias corpus and their assignments as prototypically male or female~\cite{ZWYOC18}.
The analysis shows that the Billion Word corpus contains a significant skew with respect to gender: (1) male pronouns occur three times more than female pronouns and (2) male pronouns co-occur more frequently with occupation words, irrespective of whether they are prototypically male or female.

\subsection{Geometry of Gender}
\label{sec:pca}
Next, we analyze the gender subspace in ELMo.
We first sample 400 sentences with at least one gendered word (e.g., \textit{he} or \textit{she} from the OntoNotes 5.0 dataset~\cite{weischedel2012ontonotes} and generate the corresponding gender-swapped variants (changing \textit{he} to \textit{she} and vice-versa).
We then calculate the difference of ELMo embeddings between occupation words in corresponding sentences and conduct principal component analysis for all pairs of sentences.
Figure~\ref{fig:pca} shows there are two principal components for gender in ELMo, in contrast to GloVe which only has one~\cite{BCWS16}. The two principal components in ELMo seem to represent the gender from the contextual information (Contextual Gender) as well as the gender embedded in the word itself (Occupational Gender).

To visualize the gender subspace, we pick a few sentence pairs from WinoBias~\cite{ZWYOC18}.
Each sentence in the corpus contains one gendered pronoun and two occupation words, such as
``The developer corrected the secretary because she made a mistake'' and also the same sentence with the opposite pronoun (he). 
In Figure~\ref{fig:pca} on the right, we project the ELMo embeddings of occupation words that are co-referent with the pronoun (e.g. \textit{secretary} in the above example) for when the pronoun is male (blue dots) and female (orange dots) on the two principal components from the PCA analysis.
Qualitatively, we can see the first component separates male and female contexts while the second component groups male related words such as \textit{lawyer} and \textit{developer} and female related words such as \textit{cashier} and \textit{nurse}.

\subsection{Unequal Treatment of Gender}
\label{sec:classifier}
To test how ELMo embeds gender information in contextualized word embeddings, we train a classifier to predict the gender of entities from occupation words in the same sentence.
We collect sentences containing gendered words (e.g., \word{he}-\word{she}, \word{father}-\word{mother}) and occupation words (e.g., \word{doctor})\footnote{We use the list collected in~\cite{ZWYOC18}} from the OntoNotes 5.0 corpus~\cite{weischedel2012ontonotes}, where
we treat occupation words as a mention to an entity, and the gender of that entity is taken to the gender of a co-referring gendered word, if one exists.
For example, in the sentence ``the engineer went back to her home,'' we take \word{engineer} to be a female mention.
Then we split all such instances into training and test, with 539 and 62 instances, respectively and augment these sentences by swapping all the gendered words with words of the opposite gender such that the numbers of male and female entities are balanced.

We first test if ELMo embedding vectors carry gender information. We train an SVM classifier with an RBF kernel\footnote{We use the $\nu$-SVC formulation and tune the hyper-parameter $\nu$~\cite{chang2011libsvm} in the range of $[0.1, 1]$ with a step 0.1.} to predict the gender of a mention (i.e., an occupation word) based on its ELMo embedding.
On development data, this classifier achieves 95.1\% and 80.6\% accuracy on sentences where the true gender was male and female respectively.
For both male and female contexts, the accuracy is much larger than 50\%, demonstrating that ELMo does propagate gender information to other words.
However, male information is more than 14\% more accurately represented in ELMo than female information, showing that ELMo propagates the information unequally for male and female entities.

\begin{table*}[!t]
\small
\centering
\begin{adjustbox}{max width=\textwidth}
    \begin{tabular}{|c|c|c|c|c||c|c|c|c||c|c|c|c|}
        \hline
          \multirow{2}{*}{ Embeddings} & \multirow{2}{*}{Data Augmentation} & \multicolumn{2}{c|}{Neutralization} & \multirow{2}{*}{OntoNotes} & \multicolumn{4}{c|}{Semantics Only} & \multicolumn{4}{c|}{w/ Syntactic Cues}
        \\
          \cline{3-4} \cline{6-13}

          & &  GloVe & ELMo & & Pro. & Anti. & Avg. & $\mid$ Diff $\mid$&  Pro. & Anti.& Avg. & $\mid$ Diff $\mid$\\
          \hline
          GloVe & & & &67.7 & 76.0 & 49.4 & 62.7 & 26.6* & 88.7 & 75.2 & 82.0 & 13.5* \\
           GloVe &\checkmark & & & 65.8 & 63.9 & 62.8 & 63.4 & 1.1 & 81.3 & 83.4 & 82.4 & 2.1 \\
          \hline
           GloVe+ELMo & & & &72.7 & 79.1 &49.5& 64.3 & 29.6*
           & 93.0 &85.9  & 89.5 &  7.1*\\

           GloVe+ELMo & \checkmark &  & & 71.0 & 65.9 & 64.9 &  65.4 &  1.0 & 87.8  & 88.9  & 88.4  & 1.2 \\
        \hline
        GloVe+ELMo & & \checkmark & & 71.0 & 72.6 & 57.8& 64.9 & 14.3* & 90.2 & 88.6 & 89.4 & 1.6 \\
        GloVe+ELMo & & \checkmark & \checkmark & 71.1 & 71.7 & 60.6 & 66.2 & 11.1* & 90.3 & 89.2 & 89.8 & 1.1 \\

           \hline
    \end{tabular}
    \end{adjustbox}
    \caption{F1 on OntoNotes and WinoBias development sets. WinoBias dataset is split Semantics Only and w/ Syntactic Cues subsets. ELMo improves the performance on the OntoNotes dataset by 5\% but shows stronger bias on the  WinoBias dataset. Avg. stands for averaged F1 score on the pro- and anti-stereotype subsets while ``Diff.'' is the absolute difference between these two subsets. *~indicates the difference between pro/anti stereotypical conditions is significant ($p < .05$) under an approximate randomized test~\cite{graham2014randomized}. Mitigating bias by data augmentation reduces all the bias from the coreference model to a neglect level. However, the neutralizing ELMo approach only mitigates bias when there are other strong learning signals for the task.}
    \label{tab:coref_res}
    \vspace{-10pt}
\end{table*}

\section{Bias in Coreference Resolution}
In this section, we establish that coreference systems that depend on ELMo embeddings exhibit significant gender bias.
Then we evaluate two simple methods for removing the bias from the systems and show that the bias can largely be reduced.

\subsection{Setup} We evaluate bias with respect to the WinoBias dataset~\cite{ZWYOC18}, a benchmark of paired male and female coreference resolution examples following the Winograd format~\cite{hirst,RahmanNg12c,peng2015solving}.
It contains two different subsets, pro-stereotype, where pronouns are associated with occupations predominately associated with the gender of the pronoun, or anti-stereotype, when the opposite relation is true.
Each subset consists of two types of sentences: one that requires semantic understanding of the sentence to make coreference resolution (Semantics Only) and another that relies on syntactic cues (w/ Syntactic Cues).
Gender bias is measured by taking the difference of the performance in pro- and anti-stereotypical subsets.
Previous work~\cite{ZWYOC18} evaluated the systems based on GloVe embeddings but here we evaluate a state-of-the-art system that trained on the OntoNotes corpus with ELMo embeddings~\cite{lee2018higher}.

\subsection{Bias Mitigation Methods}
Next, we describe two methods for mitigating bias in ELMo for the purpose of coreference resolution: (1) a train-time data augmentation approach and (2) a test-time neutralization approach.

\paragraph{Data Augmentation}
\label{sec:debias}
\citet{ZWYOC18} propose a method to reduce gender bias in coreference resolution by augmenting the training corpus for this task.
Data augmentation is performed by replacing gender revealing entities in the OntoNotes dataset with words indicating the opposite gender and then training on the union of the original data and this swapped data.
In addition, they find it useful to also mitigate bias in supporting resources and therefore replace standard GloVe embeddings with bias mitigated word embeddings from ~\newcite{BCWS16}.
We evaluate the performance of both aspects of this approach.

\paragraph{Neutralization}
We also investigate an approach to mitigate bias induced by ELMo embeddings without retraining the coreference model.
Instead of augmenting training corpus by swapping gender words, we generate a gender-swapped version of the test instances.
We then apply ELMo to obtain contextualized word representations of the original and the gender-swapped sentences and use their average as the final representations.

\subsection{Results}
Table~\ref{tab:coref_res} summarizes our results on WinoBias.
\paragraph{ELMo Bias Transfers to Coreference}
Row 3 in Table~\ref{tab:coref_res} summarizes performance of the ELMo based coreference system on WinoBias.
While ELMo helps to boost the coreference resolution F1 score (OntoNotes) it also propagates bias to the task.
It exhibits large differences between pro- and anti-stereotyped sets ($\mid$Diff$\mid$) on both semantic and syntactic examples in WinoBias.

\paragraph{Bias Mitigation}
Rows 4-6 in Table~\ref{tab:coref_res} summarize the effectiveness of the two bias mitigation approaches we consider.
Data augmentation is largely effective at mitigating bias in the coreference resolution system with ELMo (reducing $\mid$Diff $\mid$ to insignificant levels) but requires retraining the system.
Neutralization is less effective than augmentation and cannot fully remove gender bias on the Semantics Only portion of WinoBias, indicating it is effective only for simpler cases.
This observation is consistent with~\newcite{Gonen2019LipstickOA2}, where they show that entirely removing bias from an embedding is difficult and depends on the manner, by which one measures the bias.

\section{Conclusion and Future Work}
\label{sec:conclusion}
Like word embedding models, contextualized word embeddings inherit implicit gender bias.
We analyzed gender bias in ELMo, showing that the corpus it is trained on has significant gender skew and that ELMo is sensitive to gender, but unequally so for male and female entities.
We also showed this bias transfers to downstream tasks, such as coreference resolution, and explored two bias mitigation strategies: 1) data augmentation and 2) neutralizing embeddings, effectively eliminating the bias from ELMo in a state-of-the-art system.
With increasing adoption of contextualized embeddings to get better results on core NLP tasks, e.g. BERT~\cite{devlin2018bert},
we must be careful how such unsupervised methods perpetuate bias to downstream applications and our work forms the basis of evaluating and mitigating such bias.

\section*{Acknowledgement}
This work was supported in part by National Science Foundation Grant IIS-1760523. RC was supported by a Facebook Fellowship. We also acknowledge partial support from the Institute of the the Humanities and Global Cultures at the University of Virginia. We thank all reviewers for their comments.

\bibliographystyle{acl_natbib}
\bibliography{naaclhlt2019}

\end{document}